# Unseen Target Stance Detection with Adversarial Domain Generalization

Zhen Wang
School of Computer Science and Technology
Heilongjiang University
Harbin, China
wangzhnlp@163.com

Qiansheng Wang
School of Computer Science and Technology
Heilongjiang University
Harbin, China
chncwang@gmail.com

Chengguo Lv*
School of Computer Science and Technology
Heilongjiang University
Harbin, China
2004085@hlju.edu.cn

Xue Cao
School of Computer Science and Technology
Heilongjiang University
Harbin, China
caoxue4nlp@163.com

Guohong Fu*
Institute of Artificial Intelligence
Soochow University
Soochow, China
ghfu@suda.edu.cn

*Abstract*—Although stance detection has made great progress in the past few years, it is still facing the problem of unseen targets. In this study, we investigate the domain difference between targets and thus incorporate attention-based conditional encoding with adversarial domain generalization to perform unseen target stance detection. Experimental results show that our approach achieves new state-of-the-art performance on the SemEval-2016 dataset, demonstrating the importance of domain difference between targets in unseen target stance detection.

## I. Introduction

Stance detection, also known as stance classification, or stance identification, aims to identify the stance of a given sentence or text towards a target. As shown in Fig.1, stance detection is actually a ternary classification task and typical types of stance include favor, against, and neutral. Since social media sites like Twitter and Weibo, contain many user comments with rich stances towards specific targets, stance detection is of great value to many social media processing applications such as public opinion analysis and user analysis.

**Sentence:** *True equality allows all to be born.*

**Target:** *Legalization of Abortion*

**Stance:** *Against*

Fig. 1. An example of stance detection

Previous work [1], [2], [3] mainly focuses on training and testing on the same targets. In open applications, however, it is impossible to build a labeled or even unlabeled dataset containing all possible targets. Hence, in this paper, we focus on unseen stance detection in which targets in the test dataset are unseen in the training dataset. For example, the test set may contain the target "Hillary", while the training set only contains "Atheism" and "Feminist Movement".

To deeply exploit the relationship between the sentence and its target, Augenstein et al. [4] proposed a bi-directional conditional encoding model. It first uses Long Short Term Memory (LSTM) [5] to encode the target, and then encodes the sentence with another LSTM conditioned on the target encoding. Besides, this process is carried out in both directions. Its performance in unseen targets stance detection significantly exceeds the baseline model.

We notice that, however, the distribution of data varies among different targets. For example, the word "god" is frequently used in sentences with the target "Atheism", while "carbon" is frequently used in sentences with the target "Climate Change is a Real Concern". Based on this observation, it is arguable that conditional encoding may generalize well to unseen targets for it may overuse target specific features.

Naturally, we regard each target as a domain and our model learned labeled samples from multi-source domains during the training stage, but both samples from source domains and target domains themselves are unseen during training, which is commonly called the domain generalization problem [6].

If it cannot be distinguished which source domain representations belong to, it is supposed to generalize to unseen target domains. Hence, in this paper, we introduce adversarial domain generalization to encourage the model to learn representations that can generalize across multiple source domains. Specifically, we first obtain sentence representations through an attention-based bi-directional conditional encoding LSTM, and then implement sentence-level adversarial training to keep the representations domain-invariant. Experiments show that the proposed models with domain-invariant representations outperform their baselines. To sum up, our contributions are as follows:

- To the best of our knowledge, this work may be the first attempt to investigate the domain difference

problem for unseen target stance detection proposed by Augenstein et al. [4].
- Furthermore, we incorporate attention-based conditional encoding with adversarial domain generalization to perform unseen target stance detection.
- Finally, we conduct experiments for unseen target stance detection over the twitter stance detection dataset [7]. The experimental results show that our model outperforms the corresponding strong baselines.

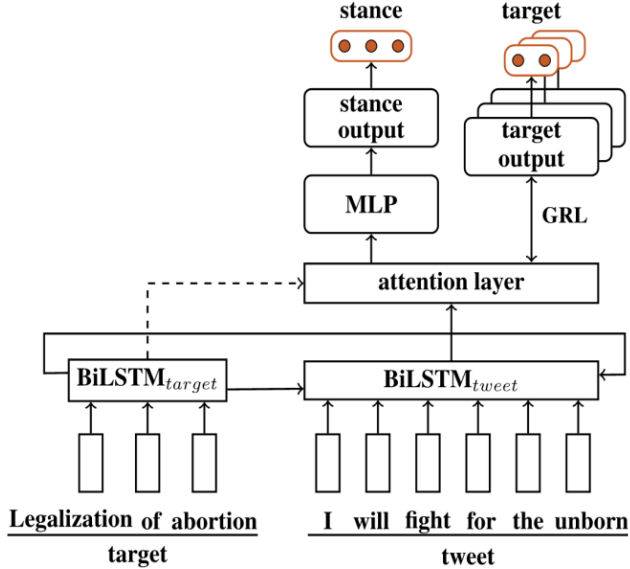

Fig. 2. Adversarail domain generalization for conditional encoding

## II. RELATED WORK

### A. Stance Detection

Mohammad et al. [7] built a Twitter dataset for stance detection. This dataset has greatly promoted the research on stance detection. Some work focuses on task settings where targets are the same between the training and test datasets [1], [2], [3], while other research explores transfer learning from one domain to another [8], [9], [10]. In the present study, we focus on transfer learning from multiple source domains to unseen target domains, which is more challenging because the model cannot learn features directly from target domains during training.

Augenstein et al. [4] proposed to deal with unseen target stance detection using conditional encoding. Following this study, we introduce adversarial domain generalization to improve the generalization capability of the conditional encoding model when classifying sentences with unseen targets. As far as we know, we are the first to follow this task.

### B. Adversarial Training

Adversarial training for domain adaptation, first proposed by Ganin et al. [11], has been widely applied in NLP tasks, including sentence classification, sequence tagging, and text generation. Li et al. [12] propose an Adversarial Memory Network (AMN) to offer direct visualization of models in cross-domain sentiment classification, which outperforms state-of-the-art methods at the time. Chen et al. [13] apply adversarial loss and domain discriminators to specific shared models using RNNs for Chinese word segmentation (CWS). Their experiments show that joint learning on multiple corpora yields a significant improvement compared to learning separately. Yang et al. [14] introduce adversarial training for Chinese named entity recognition with crowd annotations to make full use of the noisy sequence labels from multiple annotators. They show that their system achieves better performance than the baseline systems. Li et al. [15] propose a conditional sequence generative adversarial training for dialogue generation.

Xu et al. [10] first introduce adversarial learning for stance detection to tackle the problem where there is limited labeled data in the target domain, but sufficient labeled data in the source domain, and experiment results show that their model outperforms their best baseline.

Although the aforementioned as mentioned above methods can learn both domain invariant and specific features are required that target domains are available during the training stage. To remedy this shortcoming, in the present study we focus on unseen stance detection under the setting that multiple source domains labeled data are available, yet target domains remain unseen during the training stage.

### C. Domain Generalization

While domain adaptation is widely explored in both CV and NLP fields [16], [17], [18], domain generalization is still at its early stage, especially in the NLP field. Early research mainly uses all information from the training domains or datasets to learn a shared invariant representation. Ghifary et al. [19] present a new feature learning algorithm, namely the Multi-Task Autoencoder (MTAE), that can provide good generalization performance for cross-domain object recognition. Muandet et al. [20] propose a kernel-based optimization algorithm named Domain-Invariant Component Analysis (DICA) that can learn an invariant transformation by minimizing the dissimilarity. However, these approaches fail to learn shared features among different source domains. Hence, recent research [21], [22] introduces adversarial learning to encourage shared parameters to obtain domain invariant features. In this paper, we follow the latter method to tackle the unseen stance detection problem.

We consider domain generalization as an important problem in the NLP field, based on the observation that in a wide range of tasks, models trained in specific domains could be used to label corpora from unseen domains in downstream tasks. For example, the CoreNLP tools [23] are widely used to label corpora from various unseen domains that are inevitably different from source domains [24], [25]. However, only a handful of NLP work has been reported on domain generalization [22], [26], [27]. Fried et al. [27] incorporate word embeddings and BERT [28] to improve the performance of Neural Constituency Parsers in all source domains, and thus exploit structured output prediction of output trees to better generalize to out-of-domain corpora. Marzinotto et al. [22] apply adversarial training to make semantic parsing robust on both in-domain data and out-of-domain data. Different from these studies [29], [30], [31], the present study is focused on domain generalization on the sentence classification problem.

## III. THE BASELINE MODEL

### A. Conditional Encoding

To deeply exploit the relationship between a sentence and its target, we employ conditional encoding to obtain hidden states.

First, we use the LSTM to encode the target:

$$[\vec{h}_i^{target} \vec{c}_i^{target}] = \overrightarrow{LSTM}_{target}(x_i^{target}, \vec{h}_{i-1}^{target}, \vec{c}_{i-1}^{target}) \quad (1)$$

where $\vec{h}_i^{target}, \vec{c}_i^{target}$ are the hidden vector and the cell vector of the LSTM at the time i, respectively. $x_i^{target}$ is the word vector of the target at the time i, and → means forward encoding.

Then, we pass the last cell vector of $\overrightarrow{LSTM}_{target}$ to the sentence LSTM, obtaining the conditional encoding of the sentence.:

$$[\vec{h}_1^{senten} \vec{c}_1^{senten}] = \overrightarrow{LSTM}_{senten}(x_1^{senten}, \vec{h}_M^{target}, \vec{c}_M^{target}) \quad (2)$$

$$[\vec{h}_j^{senten} \vec{c}_j^{senten}] = \overrightarrow{LSTM}_{senten}(x_j^{senten}, \vec{h}_{j-1}^{senten}, \vec{c}_{j-1}^{senten}) \quad (3)$$

Where M is the length of a target.

Furthermore, to enable the model to capture both past and future contexts, we use BiLSTM [32]. Similarly, the reverse encoding process is as follows:

$$[\overleftarrow{h}_i^{target} \overleftarrow{c}_i^{target}] = \overleftarrow{LSTM}_{target}(x_i^{target}, \overleftarrow{h}_{i+1}^{target}, \overleftarrow{c}_{i+1}^{target}) \quad (4)$$

$$[\overleftarrow{h}_N^{senten} \overleftarrow{c}_N^{senten}] = \overleftarrow{LSTM}_{senten}(x_N^{senten}, \overleftarrow{h}_1^{target}, \overleftarrow{c}_1^{target}) \quad (5)$$

$$[\overleftarrow{h}_j^{senten} \overleftarrow{c}_j^{senten}] = \overleftarrow{LSTM}_{senten}(x_j^{senten}, \overleftarrow{h}_{j+1}^{senten}, \overleftarrow{c}_{j+1}^{senten}) \quad (6)$$

where N is the length of a sentence.

Finally, we concatenate the hidden vectors of the sentence from both directions:

$$h_i^{senten} = [\vec{h}_i^{senten}; \overleftarrow{h}_i^{senten}] \quad (7)$$

where $h_i^{senten}$ is the hidden vector of the LSTM at the time i, and ; means concatenation of two vectors.

### B. Domain-Related Attention

Many sentences contain salient words to which the model should pay more attention. Hence, we apply the attention mechanism to extract salient words. Based on the observation that salient words vary among different domains, attention scores are computed based on the target representation and each hidden vector of the sentence. Specifically, we use additive attention [33] as follows:

$$a_i = v^T \tanh(W_{attention}[h_i^{target}; h_i^{senten}]) \quad (8)$$

$$\alpha_i = \frac{\exp(a_i)}{\sum_j \exp(a_j)} \quad (9)$$

where $h^{target}$ is the concatenation of $\vec{h}_M^{target}$ and $\overleftarrow{h}_1^{target}$, $\alpha_i \in (0,1)$ is the attention score for each word in the sentence, and $W_{attention}$, $v$ are both trainable weight matrices.

Finally, we can obtain the sentence representation by calculating the weighted sum of sentence hidden vectors:

$$s = \sum_i \alpha_i h_i^{senten} \quad (10)$$

### C. Stance Classifier

The sentence representation $s$ is passed to a single-layer MLP with rectified linear units (ReLU) non-linearity and [34] the resulting vector is passed through a linear transformation followed by a softmax layer to obtain the stance category:

$$y = \text{ReLU}(W_{MLP}s) \quad (11)$$

$$p_{stance} = \text{softmax}(W_{stance}y) \quad (12)$$

where $p_{stance}$ is the predicted probability of the stance, and $W_{MLP}$ and $W_{stance}$ are trainable weight matrices.

## IV. ADVERSARIAL DOMAIN GENERALIZATION

We follow the setting of most domain generalization tasks that contain a large number of labeled samples from several source domains, but no labeled or unlabeled sample from the target domain during training [19], [20], [21]. Hence, the model cannot directly learn how to transfer knowledge from source domains to target domains. A common approach is to encourage the model to learn shared knowledge cross different source domains under the assumption that shared knowledge can be generalized to unseen target domains. To approach this, we apply adversarial neural networks as a regularization to the model.

Yet there are two problems to investigate: (1) Should the adversarial part be placed over the conditional or independent encoding sentence representation? (2) Can the model benefit from target-specific representations when tested on target domains? To examine the above two problems, we propose three candidate models in the following sections.

### A. Target Classifiers over Conditional Encoding

In the present study, conditional encoding is used to deeply incorporate target information into sentence representations, while target classifiers with adversarial learning are to reduce target-specific information. So it seems paradoxical to place the latter over the former. However, we argue that they will cooperate well together, as illustrated by a toy example of the sentence "I support abortion" with the target "Abortion". We hope that once the model has learned that the sentence favors the target "Abortion", it can also generalize to sentences with unseen targets (e.g., "I support Hillary" with the target "Hillary"). To achieve this generalization capability, the model should be sensitive to the target information when encoding sentences so that it could realize that "abortion" in the sentence actually refers to the target. Furthermore, such sentence representations should be highly abstract and domain-invariant. Specifically, in the above example, the stance classifier should pay attention to "support" rather than "Hillary". Hence it makes sense to place the target classifier over conditional encoding.

For each target, a corresponding binary classifier is set to predict whether the sentence belongs to the target. They are all placed over the sentence representation $s$ as Fig.2 shows, aiming to keep $s$ target-invariant. As such, each target classifier is a linear transformation of $s$, followed by a softmax layer as follows:

$$p_i^{\text{target}} = \text{softmax}(W_i^{\text{target}} s + b) \quad (13)$$

where $p_i^{\text{target}} \in \mathbb{R}^2$ is the predicted probability of whether the input vector $s$ belongs to target i, $W_i^{\text{target}}$ is a trainable matrix which is not shared among different targets, and $b \in \mathbb{R}^2$ is a trainable bias.

### B. Target Classifiers over Independent Encoding

We also implement the model with target classifiers over independent encoding as a candidate model to check whether independent encoding models could also benefit from domain invariant representations.

First, we make some modifications to the *Concat* model proposed by Augenstein et al. [4]. Different from the original Concat model, we employ BiLSTM to encode sentences and targets independently and further use max-pooling to produce two separate vectors for sentences and targets. Then we concatenate the two vectors. The detailed process is shown in formulae (14), (15), and (16).

$$s^{\text{target}} = \text{MaxPlooing}(h_0^{\text{target}}, h_1^{\text{target}}, \dots, h_M^{\text{target}}) \quad (14)$$

$$s^{\text{senten}} = \text{MaxPlooing}(h_0^{\text{senten}}, h_1^{\text{senten}}, \dots, h_M^{\text{senten}}) \quad (15)$$

$$s = [s^{\text{target}}; s^{\text{senten}}] \quad (16)$$

where $h_i^{\text{target}}, h_i^{\text{senten}}$ are max-pooled vectors of the target and the sentence, respectively.

Thus, we obtain the stance in the same way as formulae (11) and (12). For each target, s is transformed with a non-linear transformation and softmax, as illustrated above.

Finally, we place target classifiers over $s$. Here only the max-pooled vector of sentences needs to be domain-invariant, so the subsequent process of target classification is the same as formula (13).

### C. Domain Specific Representations

The stance classifier in the above models only depends on domain-invariant representations. However, it remains unclear whether domain-specific representations work. To this end, we add an additional bi-directional conditional encoding LSTM with the attention mechanism as a parallel network, feeding with the same word embeddings, but without adversarial learning. Then we obtain the representation containing both domain-invariant and domain-specific representations:

$$s = [s_{Invar}; s_{Spec}] \quad (17)$$

where $s_{Invar}$ and $s_{Spec}$ are sentence representations of domain-invariant and domain-specific models, respectively, which can be calculated using formulae (1-10).

Finally, we place the target classifier over $s$, and the subsequent process is the same as formulae (11) and (12).

### D. Training

The loss function of our models consists of two parts namely the loss function of the stance classifier and the loss function of target classifiers. We use cross-entropy for both parts. In particular, we use binary cross-entropy for each target classifier:

$$L_i = \sum_{j=0}^{N_i} [d_{ij} \log \widehat{d_{ij}} + (1 - d_{ij}) \log(1 - \widehat{d_{ij}})] \quad (18)$$

$$L_{\text{domain}} = \frac{1}{T} \sum_{i=0}^{T} L_i \quad (19)$$

where $d_{ij} \in \{0, 1\}$ denotes whether sample j belongs to domain i, $N_i$ is the number of samples from the i-th source domain, and $T$ is the number of source domains.

Finally, the loss function of the whole model is a linear combination of the stance loss and the domain loss:

$$L = L_{\text{stance}} - \lambda \cdot L_{\text{domain}} \quad (20)$$

where $\lambda$ is a positive hyperparameter, and $L_{\text{stance}}$ is the loss of the stance classifier.

To optimize L, we define $\theta$ as the set of model parameters related to the stance classifier, and $\theta'$ as the set of remaining parameters. It's worth noting that the goal of optimization is for a saddle point where both $\theta$ and $\theta'$ satisfy the condition as follows:

$$\hat{\theta} = \arg_\theta \min L(\theta, \theta') \quad (21)$$

$$\hat{\theta}' = \arg_{\theta'} \max L(\theta, \theta') \quad (22)$$

where formula (21) aims to find the $\theta$ that minimizes L, while formula (22) aims to find the $\theta'$ that maximizes $L$. Specifically, formula (21) tries to minimize the stance classifier loss, and at the same time formula (22) maximizes the target classifier loss by shared parameters of BiLSTMs and attention layers. Thus, during training, the resulting representations of BiLSTMs and attention layers will confuse target classifiers, keeping learned representations domain-invariant, and formula (22) tries to minimize the target

TABLE I. RESULTS FOR THE UNSEEN TARGET STANCE DETECTION

| Method | Stance | Dev | | | Test | | |
|---|---|---|---|---|---|---|---|
| | | P | R | F1 | P | R | F1 |
| Concat | FAVOR | 0.325 | 0.2321 | 0.2708 | 0.5 | 0.4054 | 0.4478 |
| | AGAINST | 0.6075 | 0.7673 | 0.6781 | 0.4258 | 0.6622 | 0.5183 |
| | Macro | | | 0.4745 | | | 0.4831 |
| Concat-Invar | FAVOR | 0.3313 | 0.4732 | 0.3897 | 0.4866 | 0.6149 | 0.5433 |
| | AGAINST | 0.6222 | 0.5429 | 0.5799 | 0.4223 | 0.4816 | 0.45 |
| | Macro | | | 0.4848 | | | 0.4967 |
| Augenstein et al. (2016) - BiCond | FAVOR | 0.2588 | 0.3761 | 0.3066 | 0.3033 | 0.5470 | 0.3902 |
| | AGAINST | 0.7081 | 0.5802 | 0.6378 | 0.6788 | 0.5216 | 0.5899 |
| | Macro | | | 0.4722 | | | 0.4901 |
| BCA | FAVOR | 0.3356 | 0.4464 | 0.3831 | 0.4670 | 0.5743 | 0.5152 |
| | AGAINST | 0.6122 | 0.5817 | 0.5966 | 0.4181 | 0.4950 | 0.4533 |
| | Macro | | | 0.4899 | | | 0.4842 |
| BCA-Invar-Spec | FAVOR | 0.3537 | 0.4643 | 0.4015 | 0.5191 | 0.4595 | 0.4875 |
| | AGAINST | 0.6462 | 0.5817 | 0.6122 | 0.4496 | 0.5819 | 0.5073 |
| | Macro | | | 0.5069 | | | 0.4974 |
| BCA-Invar | FAVOR | 0.375 | 0.4286 | 0.4 | 0.5029 | 0.5811 | 0.5392 |
| | AGAINST | 0.6160 | 0.6621 | 0.6382 | 0.4388 | 0.5753 | 0.4978 |
| | Macro | | | **0.5191*** | | | **0.5185*** |

(Standard deviation in parentheses)
∗ Improvements over BCA at $p < .05$ on tweet dataset

classifiers loss by its own parameters $\theta'$.

We use the standard back-propagation method to train model parameters. To jointly implement formulae (21) and (22), we introduce a gradient reverse layer (GRL) between the sentence representation layer and the target classifier, in the same way as shown in the previous work on adversarial training [10].

The GRL layer forward performs an identity transformation as follows:

$$\mathrm{GRL}(x) = x \quad (23)$$

And during backward, the gradient is simply negated.

## V. EXPERIMENTS

### A. Dataset

Our experiments are based on the twitter stance detection dataset proposed by Mohammad et al. [6], which contains a total of six targets, namely "Atheism", "Climate Change is a Real Concern", "Feminist Movement", "Legality of Abortion", "Trump" and "Hillary". Following Augenstein et al. [4], we set the first four targets as the training set, "Hillary" as the validation set, and "Trump" as the test set. Thus the training set does not include labeled or unlabeled samples from target domains and thereby target domains remain unseen in the training stage. Stance distributions of the three sets are shown in Table II.

### B. Setting

For comparison, we use the same metric approach of F1 macro-averaged over the stance FAVOR and AGAINST as in Augenstein et al. [4]. In addition, we use the same hyperparameters and settings for all candidate models as follows: (1) Word vector dimension is set to 100. (2) Hidden vector dimensions of bi-directional LSTMs are both set to 200. (3) Glove vectors [35] trained under twitter data are used and remain fixed during the training stage. (4) Dropout [36] is used after the word embedding layer, after the BiLSTM layer, and between hidden vectors in LSTMs, and all dropout rates are set to 0.1.

TABLE II. SAMPLE DISTRIBUTION OF DATASET

| Dataset | Favor | Against | None | All |
|---|---|---|---|---|
| Train | 619 | 982 | 574 | 2175 |
| Dev | 224 | 722 | 332 | 1278 |
| Test | 148 | 299 | 260 | 707 |

We use mini-batch stochastic gradient descent in the training stage, and the batch size is set to 32. To update model parameters, we use the ADAM optimizer [37], with a learning rate of 0.003 and L2 regularization of 0.01. The training process stops when the performance of the models on the validation set stops rising for 10 consecutive epochs. Then the best epoch is chosen according to the performances on the validation set.

We use the deep learning library N3LDG++[1], a powered version of N3LDG [38] to build and train our model, and our source code is available on Github[2].

### C. Comparison Models

To illustrate the performances of our proposed models, we introduce the following baseline models for comparison:

- Augenstein–BiCond: Bi-directional conditional encoding LSTM with results reported by Augenstein et al. [4].

---
[1] https://github.com/chncwang/N3LDG-plus
[2] https://github.com/Luoyufeichen/Dgnn-BiCond

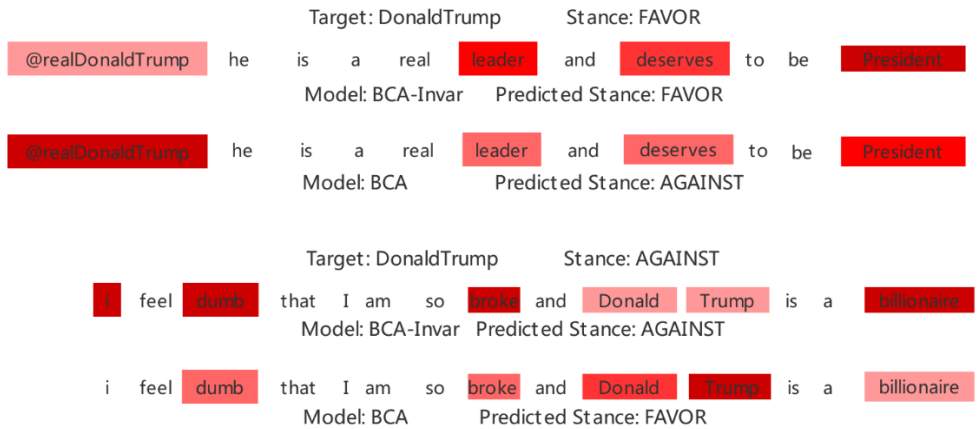

Fig. 3. Visualization of attention in the test set

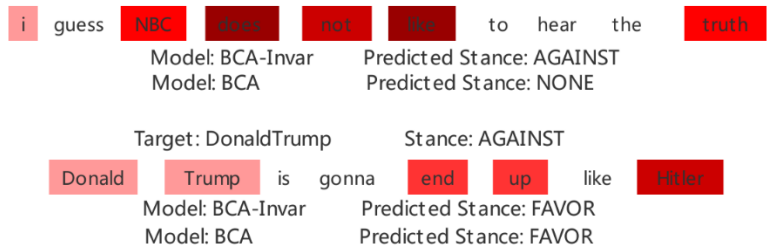

Fig. 4. Error Analysis: Visualization of attention in the test set

- BCA (BiCond with Attention): Attention-based Bi-Directional conditional encoding LSTM with results reported by us.
- Concat: Independent encoding Bi-LSTM proposed by Augenstein et al. [4] with results reported by us.

For the unseen target stance detection task, the Augenstein-BiCond model achieved the best performance at the time, and to the best of our knowledge, it was still the state-of-the-art before our work, which means a strong baseline. Besides, to compare with our proposed models fairly, we also reimplemented the BCA and Concat model with results reported by us.

Finally, our three versions of adversarial domain generalization models are marked as follows:

- BCA-Invar: the model with the adversarial classifier placed over the sentence vector of the BCA model.
- Concat-Invar: the model with the adversarial classifier placed over the concatenated sentence vectors of the Concat model.
- BCA-Invar-Spec: the BCA-Invar model with extra domain-specific representations.

### D. Results and Discussion

Table I lists results of candidate and baseline models.

**Main Results** Experimental results show that our model, BCA-Invar yielded an F1-score 0.5191 in the validation set and an F1-score of 0.5185 in the test set, exceeding Augenstein et al. [4] by 0.0469 and 0.0284, respectively. As far as we know, our model achieves the best performance in this experimental setting.

**Domain Generalization for Conditional Encoding** The adversarial domain generalization's effect is illustrated more directly by comparing BCA-Invar with BCA since they use the same hyperparameters. Our BCA model reaches 0.4899 and 0.4842 in the validation set and test set, respectively, which is comparative to the Augenstein-BiCond model (0.4722 and 0.4901). Thus, adversarial domain generalization improves the performance by F1 scores of 0.0292 in the validation set, and 0.0343 in the test set.

**Domain Generalization for Independent Encoding** The comparison between Concat and Concat-Invar shows that adversarial domain generalization can also improve the performance of the independent encoding model, though not as much as based on the conditional encoding model, with Concat-Invar exceeding Concat by F1 scores of 0.0103 in the validation set, and 0.0136 in the test set.

**Is Domain-Specific Information Useful?** After incorporating domain-specific information, BCA-Invar-Spec reaches 0.5069 in the validation set and 0.4974 in the test set, both higher than BCA, but lower than BCA-Invar. This result may suggest that when the target domain is unseen in the training stage, the model can hardly benefit from domain-specific representations, for these representations can hardly generalize to unseen domains.

### E. Visualization of Attention

To demonstrate more intuitively why BCA-Invar can better generalize to unseen domains than BCA, we select several sentences from the test dataset whose labels are predicted correctly by BCA-Invar, and then visualize the attention layer in Fig.3, where red patches highlight the words that attract more attention, and the depth of the color indicates how much attention it attracts. Specifically, Target and ground truth stance labels are displayed at the top of sentences, and stances detected by BCA and BCA-Invar are displayed at the bottom of the two sentences, respectively.

In this example, BCA-Invar predicts correctly, but BCA predicts incorrectly. It is noteworthy that BCA pays more attention to words that are more related to their targets in semantics, but the BCA-Invar prefers words related to the stance rather than target words. In detail, in the two cases, BCA highlights the words which contain "Donald" and "Trump", but BCA-Invar concentrates on salient words such as "leader", "President" and "dumb", which are more related to the stance, suggesting stronger domain generalization ability.

*F. Error Analysis*

We also analyze some cases where our model predicts the stance incorrectly. We select two cases from the test set and visualize the attention layer in Fig.4. For the first sentence, the ground truth label is "FAVOR", but our model predicts its stance as "NONE". Actually, the key point of this sentence is the relationship between "NBC" and "Trump", which should be unknown knowledge to the model. Despite better salient words BCA-Invar pays attention to, it could hardly make a correct prediction without background knowledge. Therefore, background knowledge could be quite beneficial to predict the stance correctly, and we leave it for future work.

## VI. CONCLUSIONS

To mitigate the problem that representations learned from source targets may not well generalize to unseen targets, we introduce adversarial domain generalization, achieving the best performance under the experimental setting proposed by Augenstein et al. [4]. For future work, we will explore how to exploit both labeled data from source domains and vast unlabeled data from target domains, and how to incorporate background knowledge.

## ACKNOWLEDGMENT

This work was supported by National Natural Science Foundation of China (Grant No.61672211, U1836222). Guohong Fu and Chengguo Lv are the corresponding authors.

## REFERENCES

[1] J. Du, R. Xu, Y. He, and L. Gui, "Stance classification with targetspecific neural attention networks." International Joint Conferences on Artificial Intelligence, 2017.

[2] Y. HaCohen-Kerner, Z. Ido, and R. Ya'akobov, "Stance classification of tweets using skip char ngrams," in Joint European Conference on Machine Learning and Knowledge Discovery in Databases. Springer, 2017, pp. 266–278.

[3] S. S. Mourad, D. M. Shawky, H. A. Fayed, and A. H. Badawi, "Stance detection in tweets using a majority vote classifier," in International Conference on Advanced Machine Learning Technologies and Applications. Springer, 2018, pp. 375–384.

[4] I. Augenstein, T. Rocktäschel, A. Vlachos, and K. Bontcheva, "Stance detection with bidirectional conditional encoding," arXiv preprint arXiv:1606.05464, 2016.

[5] S. Hochreiter and J. Schmidhuber, "Long short-term memory," Neural computation, vol. 9, no. 8, pp. 1735–1780, 1997.

[6] S. Ben-David, J. Blitzer, K. Crammer, A. Kulesza, F. Pereira, and J. W. Vaughan, "A theory of learning from different domains," Machine learning, vol. 79, no. 1-2, pp. 151–175, 2010.

[7] S. Mohammad, S. Kiritchenko, P. Sobhani, X. Zhu, and C. Cherry, "Semeval-2016 task 6: Detecting stance in tweets," in Proceedings of the 10th International Workshop on Semantic Evaluation (SemEval2016), 2016, pp. 31–41.

[8] K. A. Agrawal, D. Chin, and K. Chen, "Cosine siamese models for stance detection," tech. rep., Stanford University, year, Tech. Rep., 2017.

[9] M. Mohtarami, R. Baly, J. Glass, P. Nakov, L. M`arquez, and A. Moschitti, "Automatic stance detection using end-to-end memory networks," arXiv preprint arXiv:1804.07581, 2018.

[10] B. Xu, M. Mohtarami, and J. Glass, "Adversarial domain adaptation for stance detection," arXiv preprint arXiv:1902.02401, 2019.

[11] Y. Ganin, E. Ustinova, H. Ajakan, P. Germain, H. Larochelle, F. Laviolette, M. Marchand, and V. Lempitsky, "Domain-adversarial training of neural networks," The Journal of Machine Learning Research, vol. 17, no. 1, pp. 2096–2030, 2016.

[12] Z. Li, Y. Zhang, Y. Wei, Y. Wu, and Q. Yang, "End-to-end adversarial memory network for cross-domain sentiment classification." in IJCAI, 2017, pp. 2237-2243.

[13] X. Chen, Z. Shi, X. Qiu, and X. Huang, "Adversarial multicriteria learning for chinese word segmentation," arXiv preprint arXiv:1704.07556, 2017.

[14] Y. Yang, M. Zhang, W. Chen, W. Zhang, H. Wang, and M. Zhang, "Adversarial learning for chinese ner from crowd annotations," in ThirtySecond AAAI Conference on Artificial Intelligence, 2018.

[15] J. Li, W. Monroe, T. Shi, S. Jean, A. Ritter, and D. Jurafsky, "Adversarial learning for neural dialogue generation," arXiv preprint arXiv:1701.06547, 2017.

[16] E. Tzeng, J. Hoffman, K. Saenko, and T. Darrell, "Adversarial discriminative domain adaptation," in Proceedings of the IEEE Conference on Computer Vision and Pattern Recognition, 2017, pp. 7167–7176.

[17] J. Jiang and C. Zhai, "Instance weighting for domain adaptation in nlp," in Proceedings of the 45th annual meeting of the association of computational linguistics, 2007, pp. 264–271.

[18] X. Glorot, A. Bordes, and Y. Bengio, "Domain adaptation for largescale sentiment classification: A deep learning approach," 2011.

[19] M. Ghifary, W. Bastiaan Kleijn, M. Zhang, and D. Balduzzi, "Domain generalization for object recognition with multi-task autoencoders," in Proceedings of the IEEE international conference on computer vision, 2015, pp. 2551–2559.

[20] K. Muandet, D. Balduzzi, and B. Schölkopf, "Domain generalization via invariant feature representation," in International Conference on Machine Learning, 2013, pp. 10–18.

[21] J. Song, Y. Yang, Y.-Z. Song, T. Xiang, and T. M. Hospedales, "Generalizable person re-identification by domain-invariant mapping network," in Proceedings of the IEEE Conference on Computer Vision and Pattern Recognition, 2019, pp. 719-728.

[22] G. Marzinotto, G. Damnati, F. B´echet, and B. Favre, "Robust semantic parsing with adversarial learning for domain generalization," arXiv preprint arXiv:1910.06700, 2019.

[23] C. D. Manning, M. Surdeanu, J. Bauer, J. R. Finkel, S. Bethard, and D. McClosky, "The stanford corenlp natural language processing toolkit," in Proceedings of 52nd annual meeting of the association for computational linguistics: system demonstrations, 2014, pp. 55–60.

[24] A. M. Rush, S. Chopra, and J. Weston, "A neural attention model for abstractive sentence summarization," arXiv preprint arXiv:1509.00685, 2015.

[25] Q. D. Buchlak, N. Esmaili, J.-C. Leveque, F. Farrokhi, C. Bennett, M. Piccardi, and R. K. Sethi, "Machine learning applications to clinical decision support in neurosurgery: an artificial intelligence augmented systematic review," Neurosurgical review, pp. 1–19, 2019.

[26] E. Stepanov and G. Riccardi, "Towards cross-domain pdtb-style discourse parsing," in Proceedings of the 5th International Workshop on Health Text Mining and Information Analysis (Louhi), 2014, pp. 30–37.

[27] D. Fried, N. Kitaev, and D. Klein, "Cross-domain generalization of neural constituency parsers," arXiv preprint arXiv:1907.04347, 2019.

[28] J. Devlin, M.-W. Chang, K. Lee, and K. Toutanova, "Bert: Pre-training of deep bidirectional transformers for language understanding," arXiv preprint arXiv:1810.04805, 2018.

[29] J. An and P. Ai, "Deep domain adaptation model for bearing fault diagnosis with riemann metric correlation alignment," Mathematical Problems in Engineering, vol. 2020, 2020.

[30] A. Maharana and M. Bansal, "Adversarial augmentation policy search for domain and cross-lingual generalization in reading comprehension," arXiv preprint arXiv:2004.06076, 2020.

[31] J. Fu, P. Liu, Q. Zhang, and X. Huang, "Rethinking generalization of neural models: A named entity recognition case study," arXiv preprint arXiv:2001.03844, 2020.

[32] A. Graves and J. Schmidhuber, "Framewise phoneme classification with bidirectional lstm and other neural network architectures," Neural networks, vol. 18, no. 5-6, pp. 602-610, 2005.

[33] D. Bahdanau, K. Cho, and Y. Bengio, "Neural machine translation by jointly learning to align and translate," arXiv preprint arXiv:1409.0473, 2014.

[34] V. Nair and G. E. Hinton, "Rectfied linear units improve restricted boltzmann machines," in Proceedings of the 27th international conference on machine learning (ICML-10), 2010, pp. 807–814.

[35] J. Pennington, R. Socher, and C. D. Manning, "Glove: Global vectors for word representation," in Proceedings of the 2014 conference on empirical methods in natural language processing (EMNLP), 2014, pp. 1532–1543.

[36] N. Srivastava, G. Hinton, A. Krizhevsky, I. Sutskever, and R. Salakhutdinov, "Dropout: a simple way to prevent neural networks from ovefitting," The journal of machine learning research, vol. 15, no. 1, pp. 1929-1958, 2014.

[37] D. Kingma and J. Ba, "Adam: A method for stochastic optimization," Computer Science, 2014.

[38] W. Qiansheng, Y. Nan, Z. Meishan, H. Zijia, and F. Guohong, "N3ldg: A lightweight neural network library for natural language processing," Acta entiarum Naturalium Universitatis Pekinensis, 2019.